\documentclass[runningheads]{llncs}

\usepackage[table]{xcolor}
\usepackage{times}
\usepackage{epsfig}
\usepackage{graphicx}
\usepackage{amsmath}
\usepackage{amssymb}
\usepackage{cite}
\usepackage{multirow}
\usepackage{rotating}
\usepackage{array,arydshln}
\usepackage{commath}
\usepackage{verbatim}
\usepackage{pifont}
\usepackage[super]{nth}
\usepackage{booktabs}
\usepackage{subcaption}
\usepackage{gensymb}
\usepackage{slashbox}

\usepackage[width=122mm,left=12mm,paperwidth=146mm,height=193mm,top=12mm,paperheight=217mm]{geometry}
\usepackage[pagebackref=true,breaklinks=true,colorlinks,bookmarks=false]{hyperref}

\iftrue
\newcommand{\davide}[1]{\textcolor{blue}{\bf [DAVIDE: #1]}}
\newcommand{\rahul}[1]{\textcolor{green}{\bf [RAHUL: #1]}}
\newcommand{\bing}[1]{\textcolor{orange}{\bf [BING: #1]}}
\newcommand{\joe}[1]{\textcolor{magenta}{\bf [JOE: #1]}}
\newcommand{\todo}[1]{\textcolor{red}{\bf [TODO: #1]}}

\else
\newcommand{\davide}[1]{}
\newcommand{\rahul}[1]{}
\newcommand{\bing}[1]{}
\newcommand{\joe}[1]{}
\newcommand{\todo}[1]{}
\fi

\begin{document}
\pagestyle{headings}
\mainmatter
\def\ECCVSubNumber{3132}  

\title{Understanding the impact of mistakes on background regions in crowd counting} 

\titlerunning{Crowd counting mistakes on background}
\authorrunning{D. Modolo et. al}
\author{Davide Modolo, Bing Shuai, Rahul Rama Varior, Joseph Tighe}
\institute{AWS Rekognition\\
\email{\tt\small dmodolo,bshuai,rahulrv,tighej@amazon.com}}
%
\maketitle

\begin{abstract}
Every crowd counting researcher has likely observed their model output wrong positive predictions on image regions not containing any person. But how often do these mistakes happen? Are our models negatively affected by this?  
In this paper we analyze this problem in depth.
In order to understand its magnitude, we present an extensive analysis on five of the most important crowd counting datasets. We present this analysis in two parts.
First, we quantify the number of mistakes made by popular crowd counting approaches. Our results show that (i) mistakes on background are substantial and they are responsible for 18-49\% of the total error, (ii) models do not generalize well to different kinds of backgrounds and perform poorly on completely background images, and (iii) models make many more mistakes than those captured by the standard Mean Absolute Error (MAE) metric, as counting on background compensates considerably for misses on foreground. 
And second, we quantify the performance change gained by helping the model better deal with this problem. We enrich a typical crowd counting network with a segmentation branch trained to suppress background predictions. This simple addition (i) reduces background error by 10-83\%, (ii) reduces foreground error by up to 26\% and (iii) improves overall crowd counting performance up to 20\%. When compared against the literature, this simple technique achieves very competitive results on all datasets, on par with the state-of-the-art, showing the importance of tackling the background problem. 
\vspace{-3mm}
\end{abstract}

\vspace{-4mm}
\section{Introduction}
Crowd counting has attracted a lot of attention in the last few years, thanks to its applications in real-world use cases. 
%
Despite recent successes, it remains a difficult task, as models need to work well across different scenarios, from dense crowds to sparse scenes, and on any person, independently on what they are wearing or how they appear. 
One of the most important challenges is the problem of {\it scale}, which causes people on the far end of the image to appear much smaller compared to those closer to the camera. Most recent works~\cite{zhang16cvpr, sam17cvpr,sam18cvpr,sindagi17iccv,onoro16eccv,
boominathan16acm,kang18bmvc,zhang18wacv,li18cvpr,cao18eccv,ramavarior19arxiv,
liu19iccv,sindagi19iccv,zhang9iccv,xu19iccv,shi19cvpr,yan19iccv} tackle this problem by proposing new models that attempt to achieve invariance to scale variations. 

\begin{figure}[t]
\centering
    \includegraphics[width=1\textwidth]{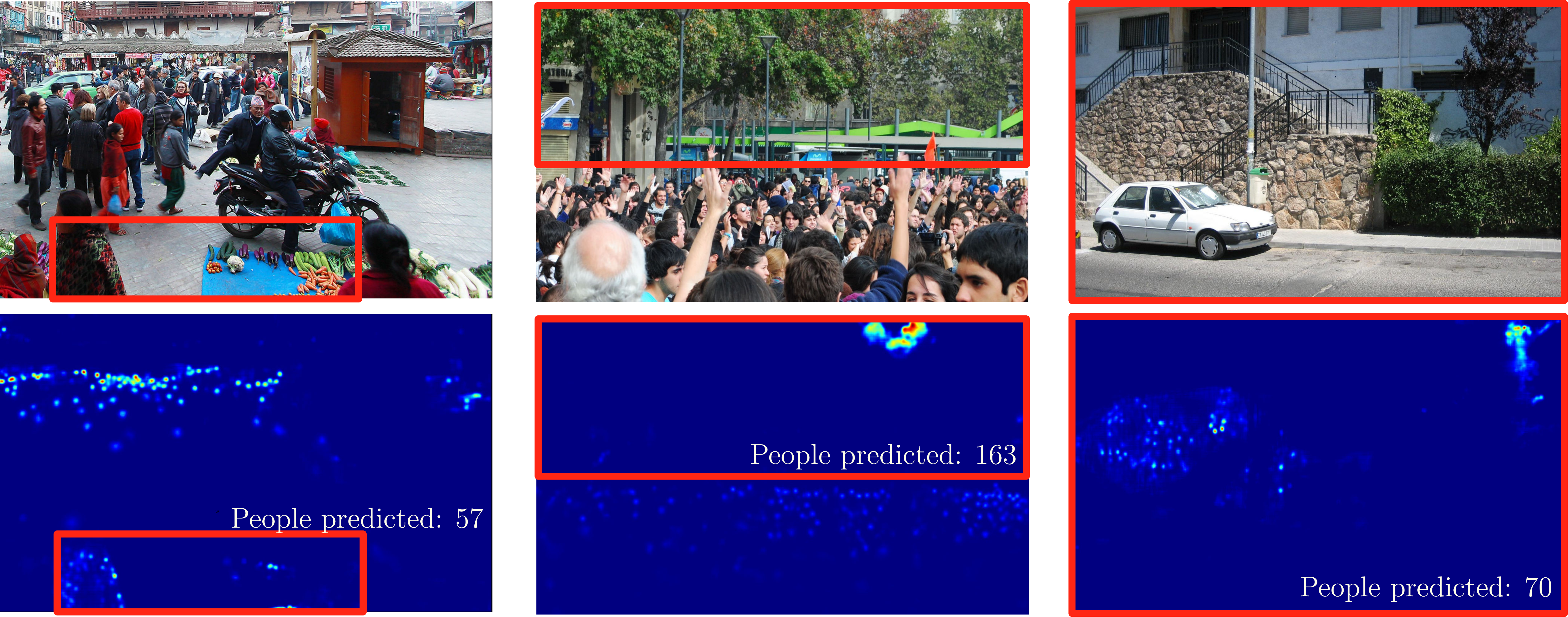}
    \caption{\small \it  Crowd counting networks output an important amount of wrong predictions on regions not containing any person, especially when these resemble crowds (e.g., foliage, roudish objects, stones, logos, etc.). \vspace{-5mm}}
    \label{fig:teaser}
\end{figure}

Instead, we investigate an orthogonal problem: errors crowd counting networks make on image regions that contain no people (i.e., the background). 
While this problem is evident in the density predictions produced by state-of-the-art crowd counting networks (fig.~\ref{fig:teaser}), it remains unexplored in the literature. Only three previous works~\cite{arteta16eccv, wan19cvpr, shi19iccv}, to the best of our knowledge, have suggested that crowd counting models should be aided to attend to foreground regions only. Here we go a step further and perform an extensive quantitative evaluation that addresses the following questions: {\it ``how much do mistakes on background affect crowd counting performance?''}
In order to answer it, we experiment with five of the most popular crowd counting datasets: Shanghai Tech (Part A \& B)~\cite{zhang16cvpr}, WorldExpo'10~\cite{zhang15cvpr}, UCF-QNRF~\cite{idrees18eccv} and GCC~\cite{wang19cvpr}. Additionally, we also experiment on ADE20k~\cite{zhou17cvprd}, a semantic segmentation dataset from which we remove people and use as 100\% background.

In the first part of this paper we focus on understanding how many mistakes are actually made on background regions.
As the concept of background is undefined for crowd counting (i.e., each person is annotated solely with a 2D point), in this work we define background as a function of a person's head size, which we estimate automatically similar to~\cite{ramavarior19arxiv}. In detail, we first enlarge each head to twice its size and then set all the pixels inside these areas as foreground and everything outside as background.
Despite these favorable conditions that relax the foreground considerably, our results show that background mistakes are responsible for 18-49\% of the total crowd counting error (depending on the dataset), On the most challenging datasets (ShanghaiTechA and UCF-QNRF), mistakes on background are almost as frequent as those on foreground (roughly 1 for every 2).
Moreover, our analysis also shows that models do not generalize well to different kinds of background. For example, a model trained on ShanghaiTechB that achieves a MAE of 5.0 on its background, achieves a much larger MAE of 18.5 on a dataset not containing any person instance (ADE20k). 
Finally, by experimenting on background and foreground independently we show that the standard crowd counting MAE computed on the full image hides a lot of mistakes, as wrong predictions on the background are used to compensate for under-predictions on the foreground. Importantly, this difference is substantial and we hope that our results will encourage the community to report MAE on background and foreground independently (e.g., the aforementioned ShanghaiTechB model that achieves a MAE of 9.1 on whole images, achieves an MAE of 5.0 on the background and a MAE of 10.7 on the foreground: $5+10.7\gg 9.1$).

In the second part on this work, we investigate how crowd counting performance changes when the network learns to tackle this problem. We propose to enhance a classic crowd counting network with a simple foreground segmentation branch used to suppress background mistakes. 
In a thorough analysis we show that this addition brings many benefits: (i) it reduces errors on background regions by 10-83\% on all datasets; (ii) it improves predictions on foreground by up to 26\%, and (iii) it increases crowd counting performance by up to 20\%. Interestingly, these improvements enable such a simple approach to achieve performance on par with the state-of-the-art methods, which use much more complex architectures. This shows the importance of addressing the background problem. 

We outline the paper as follow: in sec.~\ref{sec:rel_work} we summarize related works; in sec.~\ref{sec:background} we present our first contribution: an in-depth analysis on the impact of errors on background regions in crowd counting; in sec.~\ref{sec:decoup} we present our second contribution: an analysis on how teaching a crowd counting model about the background changes its performance; finally, in sec.~\ref{sec:concl} we present our conclusions.

\section{Related work} \label{sec:rel_work}

\noindent{\bf Crowd counting.} 
Approaches in the literature can be categorized into two high level groups: counting-by-detection~\cite{wu05iccv,wang11cvpr,rodriguez11iccv,liu18cvpr} and counting-by-regression \cite{chan08cvpr,chan09cvpr,ryan09dicta,zhang16cvpr, sam17cvpr, sam18cvpr,sindagi17iccv,onoro16eccv,
boominathan16acm,kang18bmvc,zhang18wacv,li18cvpr,cao18eccv,ramavarior19arxiv,
liu19iccv,sindagi19iccv,zhang9iccv,xu19iccv, lempitsky10nips,cheng19iccv,ma19iccv,
arteta16eccv,shi19iccv,shi19cvpr,yan19iccv}. The former group employs person/head detectors to localize and count all the instances in an image, while the latter regresses a feature representation of the image into a count number~\cite{chan08cvpr,chan09cvpr,ryan09dicta} or a density map~\cite{zhang16cvpr, sam17cvpr, sam18cvpr,sindagi17iccv,onoro16eccv,
boominathan16acm,kang18bmvc,zhang18wacv,li18cvpr,cao18eccv,ramavarior19arxiv,
liu19iccv,sindagi19iccv,zhang9iccv,xu19iccv, lempitsky10nips,cheng19iccv,ma19iccv,
arteta16eccv,shi19iccv}. 
Most of the recent approaches belong to the latter group and focus on learning new and more accurate image representations. \\

\noindent{\bf Challenges in crowd counting.} One of the most prominent challenges is the issue of {\it scale}, which causes people on the far end of the image to appear smaller than those closer to the camera. This problem originates from the perspective effect and most of the recent works in the literature have addressed this with new multi-scale models~\cite{zhang16cvpr, sam17cvpr,sam18cvpr,sindagi17iccv,onoro16eccv,
boominathan16acm,kang18bmvc,zhang18wacv,li18cvpr,cao18eccv,ramavarior19arxiv,
liu19iccv,sindagi19iccv,zhang9iccv,xu19iccv,shi19cvpr,yan19iccv}. Some works adopted multi-column architectures~\cite{zhang16cvpr,boominathan16acm,onoro16eccv,sam17cvpr,sindagi17iccv,
sam18cvpr,kang18bmvc,liu19iccv}, where each column is dedicated to a specific scale; others~\cite{zhang18wacv,li18cvpr,ramavarior19arxiv,cao18eccv,
sindagi19iccv,zhang9iccv,xu19iccv} proposed single-column architectures that learn multi-scale features within the network itself (e.g., by combining feature maps from different layers~\cite{zhang18wacv,ramavarior19arxiv,sindagi19iccv,zhang9iccv}); finally, ~\cite{shi19cvpr,yan19iccv} proposed perspective-aware networks.
On a different direction, ~\cite{lempitsky10nips,cheng19iccv,ma19iccv} focused at improving {\it spatial awareness} in counting. 
Differently from all these works, we explore yet another important problem: crowd counting networks wrongly {\it count on background} regions not containing any person's instance. While this issue was briefly mentioned in \cite{arteta16eccv,shi19iccv}, to the best of our knowledge, we are the first to quantitatively evaluate the magnitude of this problem and present an extensive analysis on how this affects crowd counting performance. \\

\noindent{\bf Reducing errors on background.}
Only three methods in the literature \cite{arteta16eccv, wan19cvpr, shi19iccv} have, to a certain extent, tried to address this problem. Arteta et al.~\cite{arteta16eccv} adopted a multi-branch architecture trained using numerous supervisions: (i) multiple point annotations from different annotators on each entity (penguin), (ii) uncertainty maps that capture the annotators agreement and (iii) depth density maps that capture the perspective change and (iv) segmentation masks for the background. Shi et al.~\cite{shi19iccv} also proposed a multi-branch architecture, but combined different maps into their final prediction: segmentation, global density and per-pixel density. Both these methods employ many cues in their approach, as they focus on achieving the best possible counting performance. Instead, in this work we employ a segmentation branch as part of our analysis, where we investigate how crowd counting performance changes as the model learns about the background. Furthermore, Wan et al.~\cite{wan19cvpr} proposed to train a semantic prior network on the ADA20k dataset and use that to re-weight the pixels with low semantic likelihood.
Differently from all these approaches that propose a new model for crowd counting, we present an analysis on this important problem and highlight several novel discoveries. 

\section{Wrong predictions on background regions} \label{sec:background}

In this section we present the first analysis on the problem of predicting counts on regions not containing any person instance. We present an extensive analysis on five of the most popular crowd counting datasets, on which we quantify the number of mistakes popular crowd counting approaches make on these regions. 




\subsection{Our baseline: CSRNet+} \label{sec:dvgg16}

In our analysis we experiment with the popular CSRNet architecture~\cite{li18cvpr}. However, in our re-implementation of this network we made few small changes to better fit it to the task of crowd counting. More specifically, we remove the \texttt{pool3} layers of VGG16 and set the dilation rates of convolution layers in the \nth{4} and \nth{5} block to be $2$ and $4$ respectively. This leads to higher-resolution features maps that are key to predicting very small people covering just a few pixels. Moreover, we also adopt a sub-pixel convolutional layer~\cite{shi16cvpr} for upsampling the predicted density map to the original input image size. From our experiments, these small changes slightly, but consistently, improve crowd counting performance over the settings of the original CSRNet (table~\ref{table:bkg_mae}). Finally, we follow the implementation details of \cite{li18cvpr} and we use the classic method proposed by Zhang et al.~\cite{zhang16cvpr} to generate ground truth density maps: we convolve each head ground truth point annotation with a fixed Gaussian kernel of $\sigma=15$ pixels.

\subsection{Datasets} \label{sec:datasets}

We experiment with five of the largest and most popular crowd counting datasets: UCF-QNRF~\cite{idrees18eccv}, Shanghai Tech (Part A \& B)~\cite{zhang16cvpr}, WorldExpo'10~\cite{zhang15cvpr} and GCC~\cite{wang19cvpr}. As these datasets capture quite different scenarios, they provide the best mix of background for this analysis. Finally, in order to understand how crowd counting models perform on general background images, we also test on a large-scale semantic segmentation dataset: ADE20K~\cite{zhou17cvprd}. 

\begin{itemize}
\item {\bf UCF-QNRF}~\cite{idrees18eccv} is one the newest crowd counting dataset and it contains large diversity both in scenes, as well as in background types. It consists of 1535 images high-resolution images from Flickr, Web Search and Hajj footage. The number of people (i.e., the count) varies from 50 to 12,000 across images.  In order to fit images as large as $6000 \times 9000$ pixels in memory, at inference we downsample these to a maximum side of $1920$ pixels. 

\item {\bf ShanghaiTech}~\cite{zhang16cvpr} consists of two parts: A and B. {\it Part A} contains 482 images of mostly crowded scenes from stadiums parades and its count averages $>$500 people per image. {\it Part B} consists of 716 images of less-crowded street scenes taken from fixed cameras and counts varying from 2 to 578.   

\item {\bf WorldExpo'10}~\cite{zhang15cvpr} contains 3980 frames from 1132 video sequences. These are split into 5 scenes and we report their average performance. The dataset is commonly evaluated by masking out images with some regions of interest (ROIs) provided by the creators, which are meant to suppress both (some) background and small non-annotated people in the far end of the image. We follow this standard procedure. 


\item {\bf GCC}~\cite{wang19cvpr} is the newest dataset and it consists of 15212 synthetic images with more than 7.5 million people. The dataset contains a large variety of computer generated scenes, from very dense to very sparse. It contains three slips: Random,  Camera and Location. We limit our analysis to the last set (as it is the most challenging one), but we compare against the literature on all three.

\item {\bf ADE20k}~\cite{zhou17cvprd} is a semantic segmentation dataset containing images picturing more diverse and challenging scenes compared to those for crowd counting. For example, these scenes range from natural to man-made and from outdoor to indoor, and they provide an excellent test use case. We evaluate on the 1468 background validation images (i.e., those that do not contain any person). 

\end{itemize}

\subsection{Metrics}
We report our results using the standard crowd counting metrics: Mean Absolute Error (MAE) and Root Mean Squared Error (MSE). In details, given the predicted count $\mathbf{C}^p$ and ground truth count $\mathbf{C}^{gt}$:
\begin{equation}
\footnotesize
\textrm{MAE} = \frac{1}{N} \sum_{i=1}^{N}|\mathbf{C}_i^p - \mathbf{C}_i^{gt}|, \; \; \; \textrm{MSE} = \sqrt{\frac{1}{N} \sum_{i=1}^{N}(\mathbf{C}_i^p - \mathbf{C}_i^{gt})^2}
\label{eq:mae}
\end{equation}
where $N$ refers to the number of test images.


In order to better analyze the behavior of crowd counting models and how they deal with background regions, we report our results on three MAE/MSE adaptations, each one evaluating the error on a particular region of the image. More specifically, we evaluate on background only, foreground only and full images. We compute these as in eq.~\ref{eq:mae}, but only count on specific regions:
\begin{equation}
\mathbf{C}_i^p = \sum_j^H \sum_k^W \mathbf{D}_i^p(j,k) \cdot \mathbf{M}^{gt}_i
\end{equation}
where  $H, W$ indicate the spatial resolution of the image, $\mathbf{D}_i^p$ is the predicted density map and $\mathbf{M}^{gt}_i$ corresponds to a ground truth mask that specifies what region to evaluate on. The computation of $\mathbf{C}_i^{gt}$ is analogous.  
For {\bf Full Image}, we set every element in $\mathbf{M}^{gt}_i$ to $1$, meaning that every pixel in the image is considered in the error estimation. Note how this is the standard case used in the crowd counting literature. For {\bf Foreground}, instead, we only set the foreground elements in $\mathbf{M}^{gt}_i$ to $1$, and the rest $0$. This estimates count error on foreground regions only and it does not penalize for false positive predictions on background. Finally, {\bf Background} has a mask complementary to Foreground (i.e., ones and zeros swapped). 
In the next section we explain how to estimate $\mathbf{M}^{gt}_i$.

\begin{figure}[t]
	\begin{center}
		\includegraphics[width=\textwidth]{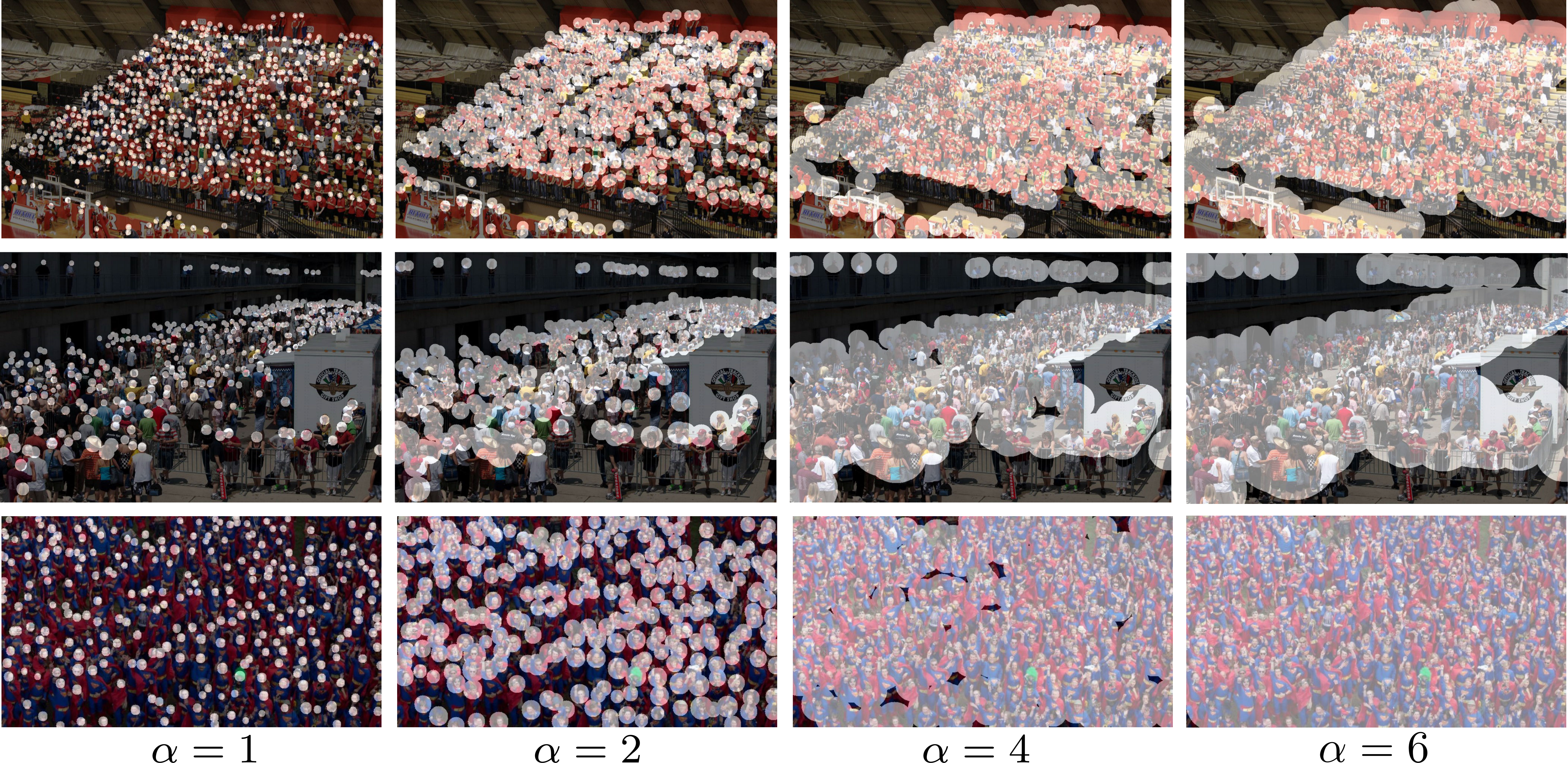}
		\vspace{-5mm} 
	\end{center}
	\caption{\small \it Three images and three foreground masks $\mathbf{M}^{gt}$ obtained by dilating the head size $d_i$ with $\alpha$ (sec.~\ref{sec:back_masks_gen}). \vspace{-2mm}}
	\label{fig:fg_mask_alpha}
\end{figure}

\subsection{Background analysis}\label{sec:back_masks_gen}
In this section we present a series of experiments that investigate if and by much crowd counting models wrongly predict people on background regions. \\

\noindent{\bf What is background in crowd counting?}
In crowd counting datasets, each person is annotated only with its head point ($x_i, y_i$), which is not sufficient to estimate good boundaries between foreground and background and to generate accurate foreground masks for evaluation. We overcome this limitation by augmenting each point annotation with a value $d_i$, corresponding to the diameter of the head. We estimate this similarly to the bounding-box technique of Rama Varior et al.~\cite{ramavarior19arxiv}: first, we run a head detector, then we associate its detections (of size $s_i$) to the annotated head points, and finally we set the size of the remaining heads (the tiny ones that the detector failed to localize) to 15 pixels, which is the common size estimate used in crowd counting. This can be summarized as $d_i = \max(s_i, 15)$. 

\begin{figure}[t]
	\begin{center}
		\includegraphics[width=0.85\textwidth]{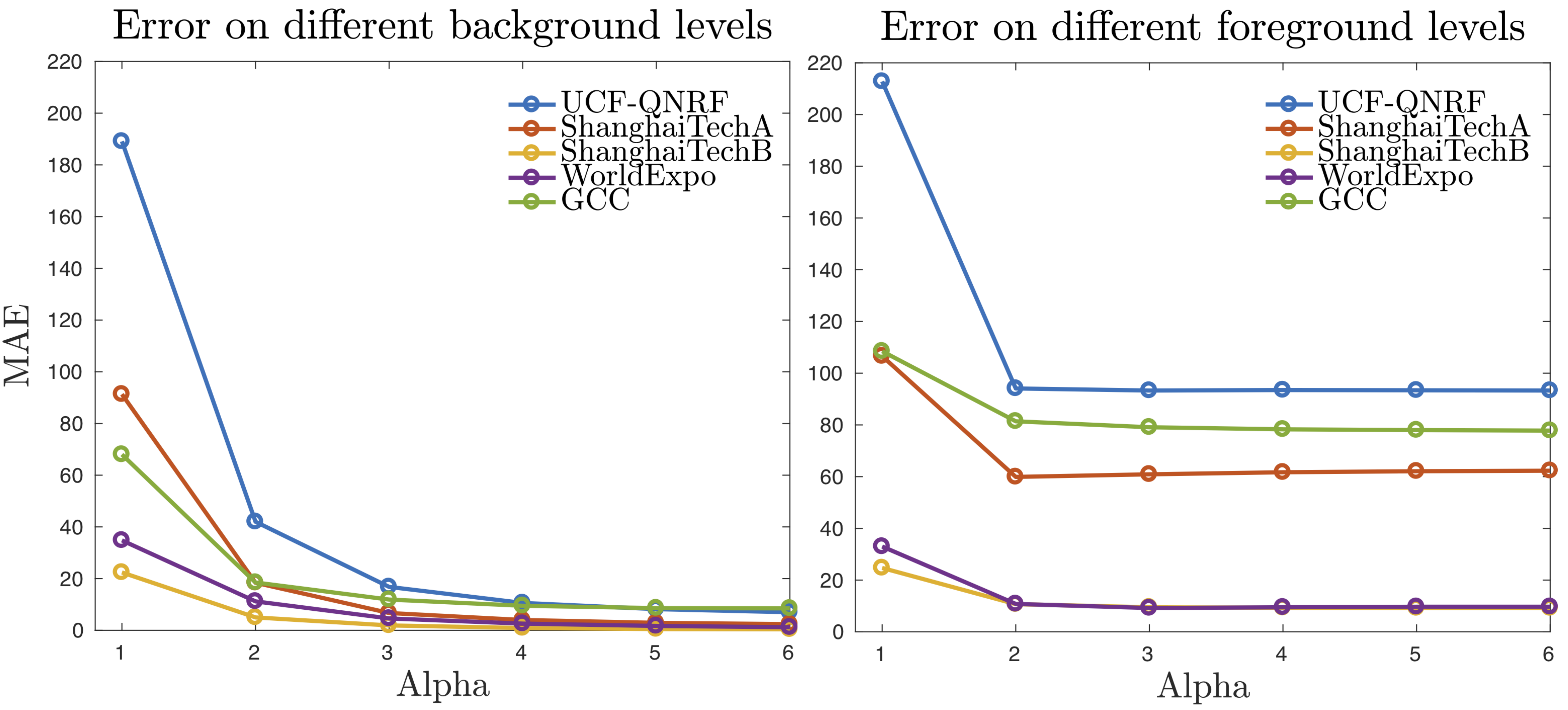}
		\vspace{-5mm} 
	\end{center}
	\caption{\small \it Errors on background and foreground regions as a function of $\alpha$, which is used to dilate each head $d_i$ and define different background/foreground boundaries (i.e., the larger $\alpha$ is, the less the amount of background in an image, fig.~\ref{fig:fg_mask_alpha} for examples).}
	\label{fig:fg_bg_curves}
\end{figure}

Next, we obtain the foreground mask $\mathbf{M}_i$ by setting all the pixels inside each head blob centered at ($x_i, y_i$) to 1. In order to understand how the performance changes with respect to the definition of foreground, we experiment with different head blob sizes by dilating the estimated head size by a factor $\alpha=[1,\dots,6]$: $d'_{i,j} = d_i\cdot\alpha_j$ (fig.~\ref{fig:fg_mask_alpha} and fig.~\ref{fig:fg_bg_curves}). 

Among these, $\alpha=1$ is the stricter definition, as each head corresponds to foreground and any non-head region is mapped to background. Under this setting, all models surprisingly achieve a very large MAE on both background and foreground. We attribute this phenomena to three factors: (i) there is uncertainty in our estimation of $s_i$, (ii) there is inconsistency in the exact location of the annotated point (i.e., sometimes the point lies on the chin of a person, other times on the forehead, etc.) and, more importantly, (iii) crowd counting models are good at counting, but less accurate at localizing each individual person: they tend to output density predictions that are less peaky than the Gaussian kernels used to convolve each point during training, resulting sometimes in predictions just outside a head region. 

For all the other values of $\alpha$, the performance is much more consistent: while foreground MAE increases slowly, background MAE continues to decrease as the background shrinks. From these results we can see that $\alpha=2$ (fig.~\ref{fig:fg_mask_alpha} mid) is a good choice to define the foreground/background boundary, as it provides a good trade-off between a too strict foreground (causing the issues mentioned above) and a too relax foreground (causing important background regions to be considered as foreground). In the remaining of the paper we present experiments using this definition. \\

\begin{table}[t]
\centering
\begin{tabular}{l | c |  c c : c c :  c }
 \toprule
     \multirow{2}{*}{Model} & Train \& Test & \multicolumn{2}{c}{Background} & \multicolumn{2}{c}{Foreground} & Full Image\\
  & dataset & \% Surface & MAE & \% Surface & MAE & MAE\\
\midrule
CSRNet+ & ShanghaiTechA & 39\% & 18.4 & 61\% & 60.0 & 64.9\\
CSRNet+ & ShanghaiTechB &  76\% & 5.0 & 24\% & 10.7 & 9.1\\
CSRNet+& UCF-QNRF &  51\% & 42.0 & 49\% & 94.1 & 95.1\\
CSRNet+ & WorldExpo & 89\% & 10.2 & 11\% & 11.2  & 8.7\\
CSRNet+ & GCC &  88\% & 17.7 & 12\% & 79.9 & 81.2\\
\midrule
MCNN~\cite{zhang16cvpr} & ShanghaiTechA  & 39\% & 56.9 & 61\% &96.1 & 110\\
MCNN~\cite{zhang16cvpr} & ShanghaiTechB  & 76\% & 18.3 & 24\% & 31.5 & 26.4\\
\midrule
CSRNet~\cite{li18cvpr} & ShanghaiTechA &  39\% &37.1 & 61\% & 68.4 & 66.5\\
CSRNet~\cite{li18cvpr} & ShanghaiTechB & 76\% & 24.2 & 24\% &24.9 & 9.6\\
\midrule
SFCN~\cite{wang19cvpr} & UCF-QNRF & 51\% & 41.6 & 49\% &114& 102\\
\bottomrule
     \end{tabular}
	\caption{\small \it MAE results of different models on five crowd counting datasets, split into background, foreground and aggregated over the full image. While the community has mostly focused on reducing Foreground error, the unexplored Background error is also important and worth addressing in the future.\vspace{-2mm}}
	\label{table:bkg_mae}
\end{table}

\noindent{\bf Is background a problem for crowd counting?}
Now that we have defined what foreground and background mean in crowd counting, we investigate how many mistakes are made in these regions with respect to the full image. 
We present our results in table~\ref{table:bkg_mae}, along with the percentage of background and foreground for each dataset. 
While not as frequent as the mistakes on foreground, the errors of CSRNet+ on background regions are still substantial: on all datasets Background MAE is responsible for around 18-49\% of the total error. 
This is especially problematic on very crowded datasets, where the areas belonging to background and foreground are very similar (ShanghaiTechA and UCF-QNRF), meaning that the errors on background are almost as frequent as those on foreground (i.e., 1:2 and 1:2.4 when normalized by the surface area). 
On the much less dense datasets, results are less severe, but this is the case because ShanghaiTechB and GCC contain similar backgrounds in their train and test sets, and WorldExpo uses ROIs to suppress difficult regions (sec.~\ref{sec:datasets}).

In table~\ref{table:bkg_mae} we also report the results we obtained by running the code and models (available online) of some popular crowd counting approaches\footnote{Note: the CSRNet models the authors released online achieve slightly better performance compared to the results they report in their paper.}. Their results show similar behavior of those of our CSRNet+ baseline. Interestingly, CSRNet achieves a substantially higher MAE on foreground and background compared to CSRNet+, even though their full image MAEs are very similar. Upon investigatation we noticed that CSRNet is not particularly good at localizing people and it tends to output less peaky density maps (due to its lower resolution feature maps).\\

\noindent {\it Observation about MAE for crowd counting.} Finally, we want to highlight how MAE computed on full images is not equal to the sum of the MAEs computed over background and foreground (e.g., in first row of table~\ref{table:bkg_mae}, $64.9 \ne 60.0+ 18.4$). 
This is surprising, considering that the union of these two mutually-exclusive pixel sets are equivalent to that of the full image.
This behavior is due to the fact that MAE computed on full images uses wrong predictions on background regions to compensate for missed predictions on foreground areas. This is an important concern, especially considering the large discrepancies reported in table~\ref{table:bkg_mae}. Going forward, we hope that these results will encourage the community to report more accurate estimates than MAE computed on the whole image, like MAE split into background and foreground, or GAME~\cite{guerrero15icpria}. \\

\noindent{\bf Do models generalize to different backgrounds?} 
Here we investigate how models trained on a specific dataset generalize to different kinds of background (i.e., to other datasets). 
Results are presented in table~\ref{table:cross_datasets}. 
The best performing model (i.e., lowest background MAE) on each dataset is, except for WorldExpo, the model trained on that same dataset. This is a domain gap problem and it is substantial; for example, a very good model trained on ShanghaiTechB makes 50\% more mistakes than one trained on UCF-QNRF on the background of UCF-QNRF (62.1 vs 42.0). This problem is even more evident in the results on the ADE20k dataset, which does not contain any person: the best model trained on real data (UCF-QNRF) outputs an average count of 8.4 per image, while the worse (WorldExpo) more than 45. These are equivalent to 12,000 and 66,000 people predicted in a dataset not containing any person. This poor generalization to different backgrounds is an important limitation towards applying crowd counting techniques to real world use cases.

Finally, the results suggest that the model trained on synthetic data (GCC) is, on average, the best performing model on background. However, upon investigation we observed that this model undercounts significantly on any real image, leading to good background MAE, but terrible foreground MAE. For example, it achieves and MAE of 235 on ShanghaiTeachA, of 25 on ShanghaiTechB, of 274 on UCF-QNRF and of 46 on WorldExpo, which are significantly higher then the results presented in table~\ref{table:bkg_mae} (MAE 60, 10.7 94.1 and 11.2, respectively). These results are a bit discouraging, as they show that significant work is needed before we can use synthetic data for crowd counting.\\

\begin{table}[t]
\centering
\resizebox{0.85\columnwidth}{!}{
\begin{tabular}{>{\arraybackslash}p{2cm} | >{\centering\arraybackslash}p{1.3cm} |  >{\centering\arraybackslash}p{1.3cm} | >{\centering\arraybackslash}p{1.3cm} | >{\centering\arraybackslash}p{1.3cm} | >{\centering\arraybackslash}p{1.3cm} | >{\centering\arraybackslash}p{1.3cm}}
 \toprule
     \multirow{2}{*}{\backslashbox{Train}{Test}} & Shanghai & Shanghai & UCF  & World & \multirow{2}{*}{GCC} & \multirow{2}{*}{ADE20k} \\
     & Tech A & Tech B & QNRF  & Expo & & \\
\midrule \centering
ShanghaiTechA & \bf \cellcolor[gray]{0.9} 18.4 & 7.7 & 57.3& 6.7 & 143.2 & 27.6\\
ShanghaiTechB & 21.3  & \cellcolor[gray]{0.9}\bf 5.0 \bf & 62.1 & 9.0 & 19.9 & 18.5\\
UCF-QNRF &  20.5 & 8.8 & \cellcolor[gray]{0.9}\bf 42.0 & 19.1 & 38.6 & 8.4\\
WorldExpo	& 98.8 & 13.5 & 118.1 & \cellcolor[gray]{0.9} 11.2 & 73.6 &  45.1\\
GCC  &  24.9  & 7.9 & 45.2 & \bf 5.9 &  \bf \cellcolor[gray]{0.9} 17.7 & \bf 3.2\\
\bottomrule
     \end{tabular}}
	\caption{\small \it Background MAE for CSRNet+ across datasets. \vspace{-4mm}}
	\label{table:cross_datasets}
\end{table}

\noindent{\bf Conclusions.} 
Crowd counting models occasionally make mistakes on background regions and every researcher on this topic is likely aware of this behavior. However, this problem appears to be much more severe that what people may have anticipated: 
our analysis quantitatively showed that crowd counting models produce an important number of wrong predictions on background regions, which flactuates from 18 to 49\%, depending on the dataset. 
Moreover, our analysis also showed why these mistakes have not been clearly captured before: the MAE metric computed on the full image hides these mistakes behind underpredictions on foreground regions, fooling us in believing that crowd counting models perform better than they actually do.  
Finally, our analysis also showed that crowd counting datasets do not contain enough diversity in terms of background, which lead to poor generalization when evaluated on pure background images. Given all these discoveries, we believe that wrongly predicting people on background regions is an important issue in crowd counting and we hope that these results will inspire more works to solve this problem. 

\begin{figure}[t]
\begin{center}
   \includegraphics[width=1\textwidth]{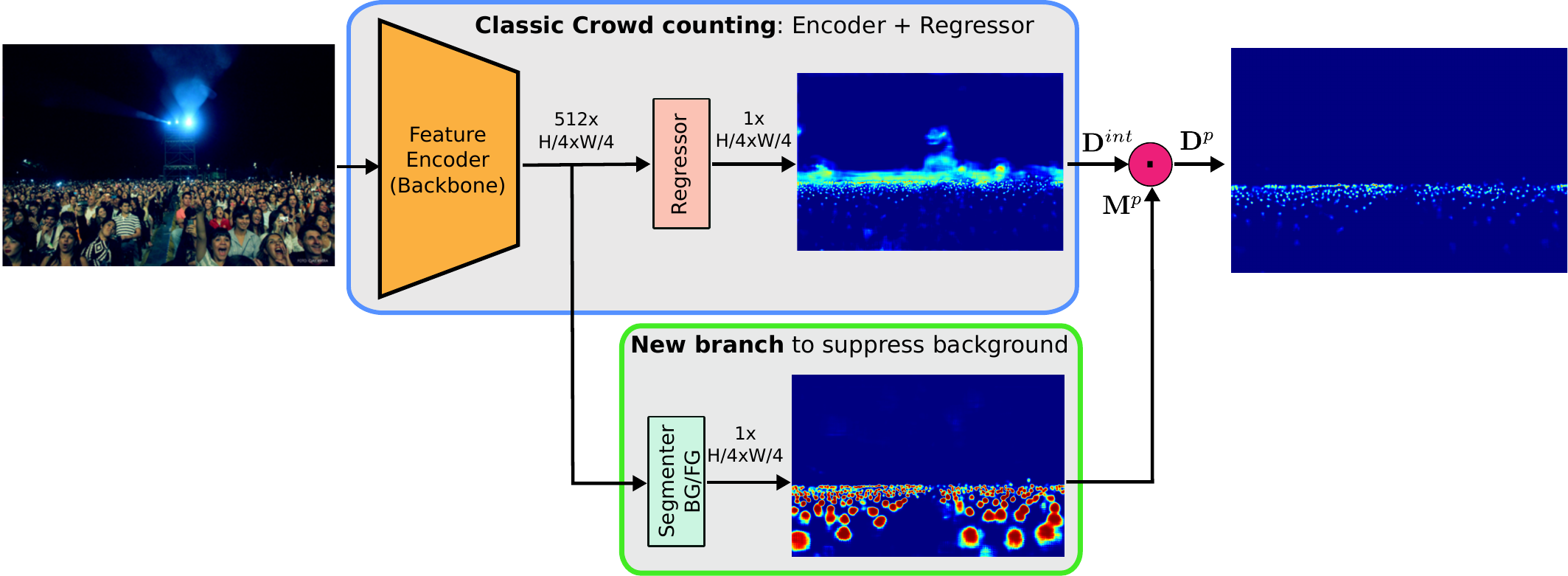}
\end{center}
\vspace{-5mm}
\caption{\small \it In the top row we show a classic crowd counting methods that consists of a  feature encoder and a count regression module. In our approach we enrich this design with a segmentation branch that is used to suppress predictions on background regions.\vspace{-5mm}} 
\label{Fig:architecture}
\end{figure}

\section{Teaching the network about background} \label{sec:decoup}
\vspace{-5mm}
In this section we present the second part of our analysis, where we investigate how crowd counting performance changes when a network learns to suppress wrong counts on background regions. Towards this, we propose a simple change to the typical crowd counting network that aims at reducing background mistakes from the final density map, leading to cleaner outputs and more accurate counting.
We propose to enrich the final regression block that maps the backbone's features to a density map for crowd counting, with a new head that is trained specifically to suppresses predictions on background regions (fig.~\ref{Fig:architecture}). We model this head as a shallow foreground/background segmentation branch that has low impact in terms of computational cost with respect to the overall model.  
This segmentation head outputs a mask that is used to modulate the density map outputted by the regression head. This  mechanism has two benefits: while the segmentation head reliably suppresses background predictions, the regression head can now better specialize on foreground patterns and improve its counting accuracy (due to being trained on gradients from foreground pixels only, eq.~\ref{eq:final_loss}).

Note that the idea of using a segmentation branch to attend to foreground regions was first introduced by Arteta et al.~\cite{arteta16eccv} to count penguins and very recently also used
by Shi et al.~\cite{shi19iccv} to count people. However, in both these works this was just a small component of a more complex design, as their goal was to achieve the best counting performance. Instead, we aim at understanding the impact of suppressing background mistakes on the final output and we are purposely making our model as simple and as specialized to this problem as possible. Moreover, differently from \cite{arteta16eccv,shi19iccv}, our work proposes a minimal change into how crowd counting models are trained and it can be easily added to most of the crowd counting approaches that focus at improving the backbone encoder. In the next paragraphs we explain how to train this simple approach.\\

\noindent {\bf Loss functions.}
Given a training image of size ($W$, $H$), we train the segmentation head with a pixel-wise binary cross entropy loss between the sigmoided predicted mask $\mathbf{M}^p$ and its corresponding ground truth $\mathbf{M}^{gt}$: 


\begin{equation}
\mathcal{L}_{bce} = \frac{1}{HW}\sum_i^H \sum_j^W -\mathbf{M}_{i,j}^{gt} \cdot log(\mathbf{M}_{i,j}^p) - (1-\mathbf{M}_{i,j}^{gt}) \cdot log(1-\mathbf{M}_{i,j}^p)
\end{equation}


Moreover, following the literatures~\cite{zhang16cvpr, sam17cvpr,sam18cvpr,sindagi17iccv,onoro16eccv,
boominathan16acm,kang18bmvc,zhang18wacv,li18cvpr,cao18eccv,ramavarior19arxiv,
liu19iccv,sindagi19iccv,zhang9iccv,xu19iccv,shi19cvpr,yan19iccv}, we train our regression head with a pixel-wise $\ell_2$ distance loss between the predicted density map $\mathbf{D}^{p}$ and its corresponding ground truth map $\mathbf{D}^{gt}$: 
\begin{equation}
 \mathcal{L} _{\ell_2}= \sum_i^H \sum_j^W \sqrt{(\mathbf{D}_{i,j}^p - \mathbf{D}_{i,j}^{gt})^2}
  \label{eq:l2}
\end{equation} 
where the predicted density map $\mathbf{D}^{p}$ is obtained by modulating the intermediate density map $\mathbf{D}^{int}$ with the predicted foreground mask $\mathbf{M}^p$ as follows: $\mathbf{D}^p = \mathbf{D}^{int} \odot \mathbf{M}^p$, in which $\odot$ represents the Hadamard operation. Importantly, note how the regression module only aggregates counting contribution for foreground regions, as the segmentation head takes the responsibility for recognizing background regions. 
%
In an end-to-end fashion (fig.~\ref{Fig:architecture}), we train our model (including the backbone) with the following dual task loss:
\begin{equation}
\mathcal{L} = \mathcal{L}_{\ell_2}(\mathbf{D}^p, \mathbf{D}^{gt}) + \lambda \mathcal{L}_{bce}(\mathbf{M}^p, \mathbf{M}^{gt})
\label{eq:final_loss}
\end{equation}
where $\lambda$ regulates the importance of the segmentation loss. From these losses one can see how separating foreground and background predictions (to regressor and segmenter, respectively) not only helps reducing mistakes on background, but it also helps the regressor becoming more accurate, as it is now entirely dedicated on counting on foreground regions only. \\

\noindent{\it Implementation details.} We use the backbone (CSRNet+) introduced in sec.~\ref{sec:dvgg16} and 3 additional fully convolutional layers for the segmentation head (an exact copy of the 3 fully convolutional layers used for regression). Moreover, we generate $\mathbf{M}^{gt}$ as explained in sec.~\ref{sec:back_masks_gen}, with $\alpha=1$. 


\subsection{Validation of our approach} \label{sec:ablation}
In this section, we experiment with this simple approach and evaluate its impact on the task of crowd counting, especially on background regions. 
We compare the CSRNet+ baseline model presented in sec.~\ref{sec:dvgg16} with the same CSRNet+, but enhanced with a segmentation branch. For simplicity, in the remaining of the paper we will refer to our approach as CSRNet+ w/BS (background suppression). First, we compare against the CSRNet+ results presented in table~\ref{table:bkg_mae} and investigate if this new branch can improve its performance. These are presented in table~\ref{table:results_regression_count}.\\

\vspace{-1mm}
\noindent {\it Background mistakes decrease  (Background MAE).}  Results validate our hypothesis that the segmentation branch can help reduce mistakes on background and show that our approach can consistently reduce these errors by an important 10-40\% on all datasets, over the baseline. Importantly, this improvement generalizes better to other background types, like those in background dataset ADE20k (table~\ref{table:res_ade20k}), where MAE always decreases, from 30\% to more than 80\%.\\

\begin{table}[t]
\centering
\resizebox{0.85\columnwidth}{!}{
\begin{tabular}{>{\centering\arraybackslash}p{2.5cm} | >{\arraybackslash}p{2.5cm} | >{\centering}p{1cm}  >{\centering\arraybackslash}p{1.3cm} | >{\centering}p{1cm}  >{\centering\arraybackslash}p{1.3cm} | >{\centering}p{1cm}  >{\centering\arraybackslash}p{1.3cm}}
 \toprule
    Train \& Test & \multirow{2}{*}{$\;$ Model} & \multicolumn{2}{c|}{Background} & \multicolumn{2}{c|}{Foreground} & \multicolumn{2}{c}{Full Image}\\
    dataset  & & \multicolumn{2}{c|}{MAE} & \multicolumn{2}{c|}{MAE} & \multicolumn{2}{c}{MAE}\\
     \midrule
     \multirow{2}{*}{ShanghaiTechA} & $\;$CSRNet+ & 18.4 & \multirow{2}{*}{$\downarrow\downarrow$19\%} & 60.0 & \multirow{2}{*}{$\downarrow$1.6\%} & 64.9 & \multirow{2}{*}{$\downarrow$3.5\%}\\
        & $\;$CSRNet+ w/BS & 14.9 &  & 58.9 &  & 62.6 & \\
        \midrule
        \multirow{2}{*}{ShanghaiTechB} & $\;$CSRNet+ & 5.0 & \multirow{2}{*}{$\downarrow\downarrow$36\%} & 10.7 & \multirow{2}{*}{$\downarrow\downarrow$26\%} & 9.1 & \multirow{2}{*}{$\downarrow\downarrow$20.1\%}\\
        & $\;$CSRNet+ w/BS & 3.2 &  & 7.9 &  & 7.2 & \\
        \midrule
        \multirow{2}{*}{UCF-QNRF} & $\;$CSRNet+ & 42.0 & \multirow{2}{*}{$\downarrow\downarrow$24\%} & 94.1 & \multirow{2}{*}{$\downarrow$9.1\%} & 95.1 & \multirow{2}{*}{$\downarrow$9.2\%}\\
        & $\;$CSRNet+ w/BS & 31.9 &  & 85.5 &  & 86.3 & \\
        \midrule
                \multirow{2}{*}{WorldExpo} & $\;$CSRNet+ & 11.2 & \multirow{2}{*}{$\downarrow$10\%} &  10.2 & \multirow{2}{*}{$\downarrow$7\%} & 8.7 & \multirow{2}{*}{$\downarrow$6.9\%}\\
        & $\;$CSRNet+ w/BS & 10.1 &  & 9.5 &  & 8.1 & \\
        \midrule
        \multirow{2}{*}{GCC} & $\;$CSRNet+ & 17.7 & \multirow{2}{*}{$\downarrow\downarrow$43\%} & 79.9 & \multirow{2}{*}{$\downarrow\downarrow$17.3\%} & 81.2 & \multirow{2}{*}{$\downarrow\downarrow$19.2\%}\\
        & $\;$CSRNet+ w/BS & 10.1 &  & 66.1 &  & 65.6 & \\
     \bottomrule
     \end{tabular}}
	\caption{\small \it We compare CSRNet+ w/o and w/ our background suppression branch (BS) on five crowd counting datasets. Adding the background suppression branch brings many benefits: (i) errors on background reduce considerably, (ii) errors on foreground also reduce, though less and (iii) the final performance is always better. \vspace{-5mm}}
	\label{table:results_regression_count}
\end{table}

\begin{table}[t]
\centering
\resizebox{0.95\columnwidth}{!}{
\begin{tabular}{>{\arraybackslash}p{2.5cm} | >{\centering}p{1cm}  >{\centering\arraybackslash}p{1cm} | >{\centering}p{1cm}  >{\centering\arraybackslash}p{1cm} | >{\centering}p{1cm}  >{\centering\arraybackslash}p{1cm} |>{\centering}p{1cm}  >{\centering\arraybackslash}p{1.1cm} | >{\centering}p{1cm}  >{\centering\arraybackslash}p{1cm}}
 \toprule
 
 \multirow{2}{*}{\backslashbox{Model}{Train}}& \multicolumn{2}{c|}{Shanghai} & \multicolumn{2}{c|}{Shanghai} & \multicolumn{2}{c|}{UCF-} & \multicolumn{2}{c|}{World} & \multicolumn{2}{c}{\multirow{2}{*}{GCC}}\\
      & \multicolumn{2}{c|}{TechA} & \multicolumn{2}{c|}{TechB} & \multicolumn{2}{c|}{QNRF} & \multicolumn{2}{c|}{Expo} & \\
     \midrule
     CSRNet+ & 27.6 & \multirow{2}{*}{$\downarrow\downarrow$29\%} & 18.5 & \multirow{2}{*}{$\downarrow\downarrow$83\%} & 8.4 & \multirow{2}{*}{$\downarrow\downarrow$38\%} & 45.1 & \multirow{2}{*}{$\downarrow\downarrow$20\%} & 3.2 &\multirow{2}{*}{$\downarrow\downarrow$75\%}\\
        CSRNet+ w/BS & 19.7 &  & 3.1 &  & 5.2 & & 36.0 & & 0.8 & \\
     \bottomrule
     \end{tabular}}
	\caption{\small \it We compare CSRNet+ w/o and w/ our background suppression branch (BS) on the pure background dataset ADE20k. Errors on background reduces substantially. \vspace{-2mm}}
	\label{table:res_ade20k}
\end{table}

\vspace{-1mm}
\noindent {\it Foreground errors decrease (Foreground MAE).} In sec.~\ref{sec:decoup}, we argued that adding the segmentation mask and using it to modulate the final output can, in theory, help the regressor better specialize on foreground (as it is lifted of the responsibility of the background) and produce more accurate predictions. Results validate this and show that in practice MAE on foreground regions always reduces, sometimes marginally (+1.6\% on Shanghai Tech A), but other times substantially (+26\% on Shanghai Tech B).\\

\vspace{-1mm}
\noindent {\it Overall performance improves (Full Image MAE).}
Finally, results also show that improving both background and foreground MAEs leads to a consistent improvement in the overall image performance, up to 20\% better. 


\begin{figure}
	\begin{center}
		\includegraphics[width=1 \textwidth]{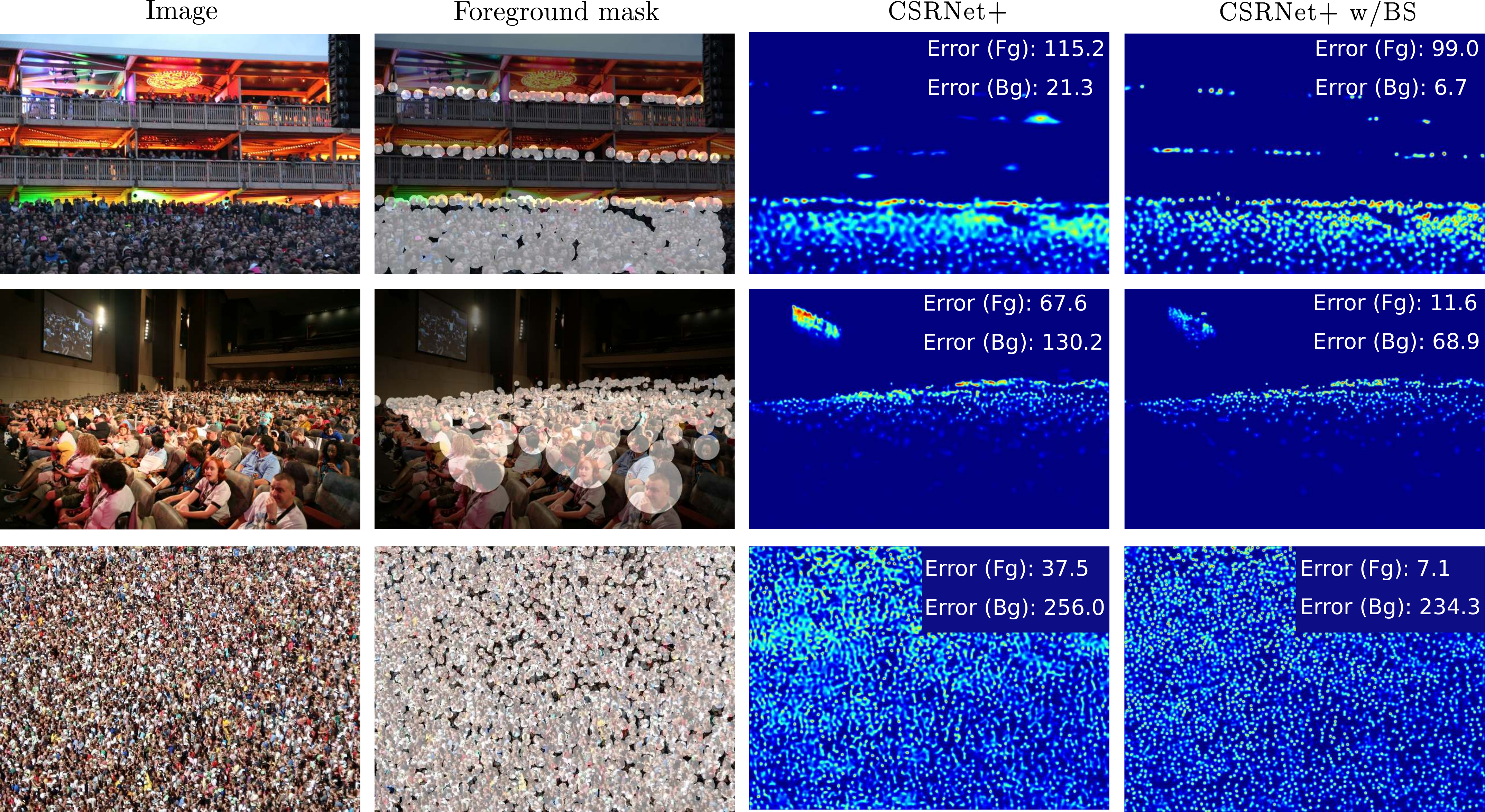}
		\vspace{-5mm}
	\caption{\small \it Enhancing CSRet+ with the ability to suppress background (w/BS) produces more accurate density maps with less errors on background regions (Error Bg) and significantly sharper foreground estimates (Error Fg).\vspace{-6mm}}
	\label{Fig:results_regression_count}
	\end{center}
\end{figure}

\begin{figure}[t]
	\centering  
	\includegraphics[width=0.9\textwidth]{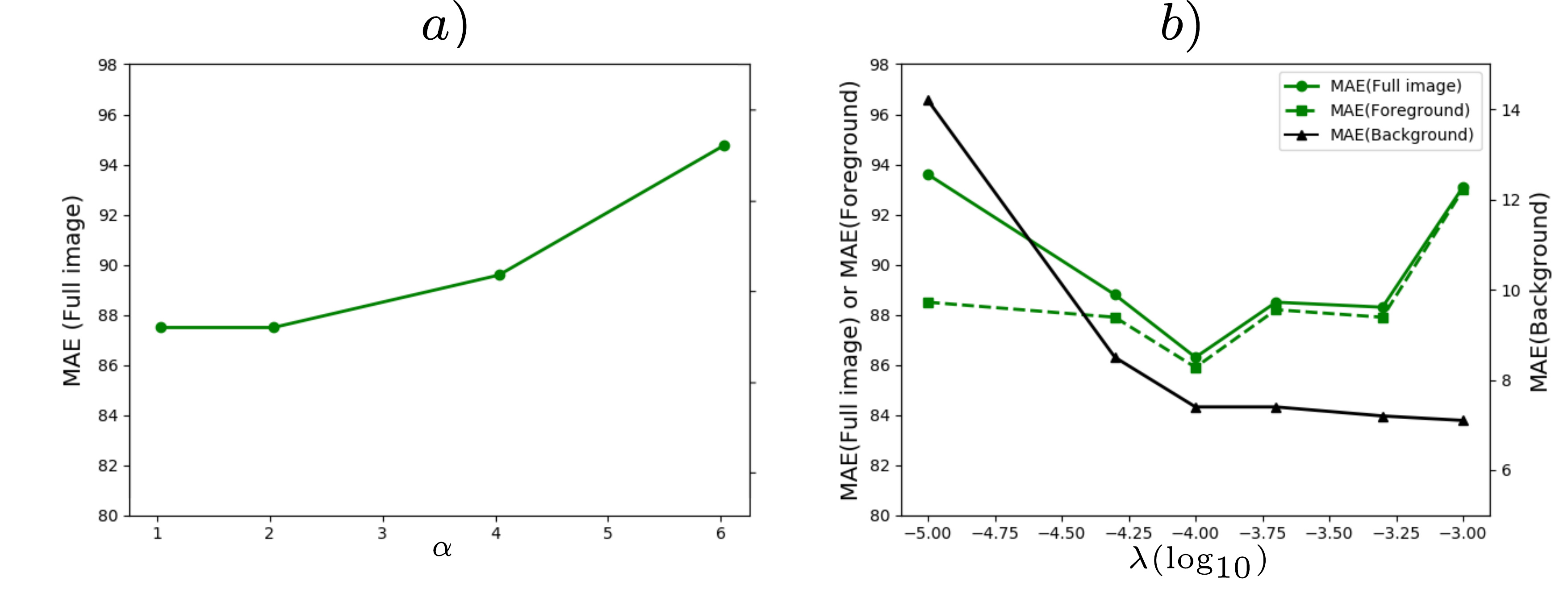}
	\vspace{-4mm}
	\caption{\it \small Results for different values of $\alpha$ (a) and $\lambda$ (b). \vspace{-2mm}}
	\label{fig:hyperparams}
\end{figure}

\subsection{Sensitivity analysis of hyper-parameters}\label{sec:params}
In this section, we evaluate some of our choices for the paraneters if CSRNet+ w/BS. We experiment on the UCF-QNRF dataset only, as it is the largest and most diverse crowd counting dataset and it is a good benchmark for this study. \\

\noindent {\bf Lambda $\lambda$.} We investigate how sensitive our model is to different values of $\lambda$ in eq.~\ref{eq:final_loss}, which regularizes the importance of the segmentation head. As shown in fig.~\ref{fig:hyperparams}{\color{red}b}, the model performs best on full images when $\lambda$ is in the range of $0.5\times10^{-4}$ to $5\times10^{-4}$. This provides a good trade-off between not using the segmentation head ($\lambda=0$) and relying on it too much ($\lambda$ too large).
Moreover, MAE on foreground is the lowest when $\lambda$ is around $10^{-4}$, while MAE on background consistently decreases as the segmentation head gets more and more importance (i.e., $\lambda$ increases). These results show the importance of training a model that is well balanced and performs well on both foreground and background regions. In our experiments, we use $\lambda$ to $10^{-4}$.\\

\noindent {\bf Training foreground mask $\mathbf{M}^{gt}$.} In sec.~\ref{sec:decoup} we chose $\alpha=1$ to generate the ground truth foreground masks used to train our segmentation head ($d_i = \max(s_i, 15)\cdot\alpha$). Our intuition for this is twofold: (i) $d$ should be at least as large as the Gaussian kernel $\sigma=15$ used to define GT maps (otherwise non-zero pixels' count would be wrongly assigned to background) and (ii) it should be as close as possible to the true size of the head ($s_i$).  To verify this hypothesis, we experiment here with different values of $\alpha$ and train with different foreground/background definitions. Results in fig.~\ref{fig:hyperparams}{\color{red}a} show that the best performance is indeed achieved by setting $\alpha$ to 1. Nevertheless, results also show that our model can deal well with more relaxed masks and still achieve great performance until the point where $d_i$ becomes too large and almost every pixel is labeled as foreground (i.e., $\alpha>4$).

\begin{table}[t]
\centering
	\resizebox{\textwidth}{!}{
		\begin{tabular}{ l | r l |  >{\centering}p{1cm}  >{\centering\arraybackslash}p{1cm} | >{\centering}p{1cm}  >{\centering\arraybackslash}p{1cm} | >{\centering}p{1cm}  >{\centering\arraybackslash}p{1cm} | >{\centering\arraybackslash}p{1.5cm} | >{\centering}p{1cm} >{\centering}p{1cm} >{\centering\arraybackslash}p{1cm}}	
			\toprule
			\multicolumn{3}{c|}{} & \multicolumn{2}{c|}{UCF-QNRF} & \multicolumn{2}{c|}{Shanghai Tech A} & \multicolumn{2}{c|}{Shanghai Tech B} & \multicolumn{1}{c|}{WorldExpo} & \multicolumn{3}{c}{GCC (MAE)}\\
			\textbf{Method} & \multicolumn{2}{c|}{\textbf{Venue \& Year}} & MAE   & MSE   & MAE   & MSE   & MAE   & MSE   & Avg MAE & Rand & Cam & Loc\\  
			\midrule
			MCNN~\cite{zhang16cvpr}    & CVPR & 2016	& 277 	& 426 	& 110.2 & 173.2 & 26.4 	& 41.3  & 11.6  & 100.9 & 110.0 & 154.8\\
			SwitchCNN~\cite{sam17cvpr}  & CVPR & 2017	& 228 	& 445 	& 90.4 	& 135.0 & 21.6 	& 33.4   &  9.4 & -& -& -\\
			CP-CNN~\cite{sindagi17iccv} & ICCV & 2017  & - 	& - 	& 73.6 	& 106.4 & 20.1 	& 30.1  &  8.9 & -& -& -\\
			SaCNN~\cite{zhang18wacv}    & WACV & 2018  & -     & -     & 86.8  & 139.2 & 16.2  & 25.8  & 8.5  & -& -& -\\
			IG-CNN~\cite{sam18cvpr}     & CVPR & 2018  & -     & -     & 72.5  & 118.2 & 13.6  & 21.1 &  11.3 & -& -& -\\
			CSRNet~\cite{li18cvpr} 	    & CVPR & 2018	& - 	& - 	& 68.2 	& 115.0 & 10.6 	& 16.0  &  8.6 & 38.2 & 61.1 & 92.2\\ 
			CL-CNN~\cite{idrees18eccv}  & ECCV & 2018  & 132 	& 191 	& - 	& - 	& - 	&  -  & -  & -& -& -\\
			SANet~\cite{cao18eccv}    & ECCV & 2018  & -  	& -	    & 67.0	& 104.5 &  8.4	&  13.6 & 8.2  & -& -& -\\
			PACNN~\cite{shi19cvpr} & CVPR & 2019 & - & - & 66.3 & 106.4 & 8.9 & 13.5 & 7.8 & -& -& -\\
			CSRNet+PACNN~\cite{shi19cvpr} & CVPR & 2019 & - & - & 62.4 & 102.0 & 7.6 & 11.8 & - & -& -& -\\
			SFCN~\cite{wang19cvpr} & CVPR & 2019 & 102.0 & 171.4 & 64.8 & 107.5 & 7.6 & 13.0 & 9.4&  {\bf 36.2} &  {\bf 56.0} & {\bf 89.3}\\
			FF~\cite{shi19iccv} & ICCV & 2019 & 93.8 & {\bf 146.5} & 65.2 & 109.4 & 7.2 & 12.2 & -& -& - & -\\
			DSSINet~\cite{liu19iccv} & ICCV & 2019 & 99.1 & 159.2 & 60.6 & 96.0 & 6.8 & 10.3 & {\bf 6.7}& -& -& -\\
	       BL~\cite{ma19iccv} & ICCV & 2019 & {\bf 88.7} & 154.8 & 62.8 & 101.8 & 7.7 & 12.7 & - & -& - & -\\		
	       MBTTBF-SCFB~\cite{sindagi19iccv}	 & ICCV & 2019 & 97.5 & 165.2 & 60.2 & 94.1 & 8.0 & 15.5 & -  & -& -& -\\
	       PGCNet~\cite{yan19iccv} & ICCV & 2019 & - & - & \bf 57.0 & \bf 86.0 & 8.8 & 13.7 & 8.1 & - & - & -\\
			CSRNet+SPANet~\cite{cheng19iccv} & ICCV & 2019 & - & - & 62.4 & 99.5 & 8.4 & 13.2 & 7.9& -& -& -\\
			SANet+SPANet~\cite{cheng19iccv} & ICCV & 2019 & - & - &  59.4 & 92.5 & {\bf 6.5} & {\bf 9.9} & 7.7& -& -& -\\
			\midrule
				CSRNet+ w/BS  & \multicolumn{2}{c|}{-}   & \cellcolor[gray]{0.9}\bf 86.3    & 	\cellcolor[gray]{0.9} \bf 153.1 	& \cellcolor[gray]{0.9} 62.6 & \cellcolor[gray]{0.9} 103.3 	&  \cellcolor[gray]{0.9} 7.2 	& \cellcolor[gray]{0.9} 11.5  & \cellcolor[gray]{0.9} 8.1 & \cellcolor[gray]{0.9} \bf 30.2 & \cellcolor[gray]{0.9} \bf 39.3 & \cellcolor[gray]{0.9} \bf 65.6\\
	 CSRNet+  w/BS (pre-trainedß) & \multicolumn{2}{c|}{-}   & -   & -	& \cellcolor[gray]{0.8} \bf 58.3 & \cellcolor[gray]{0.8} \bf 100.1 	&  \cellcolor[gray]{0.8} \bf 6.7 	& \cellcolor[gray]{0.8} \bf 10.7  & \cellcolor[gray]{0.8} \bf 7.9 & \cellcolor[gray]{0.8} 32.6 & \cellcolor[gray]{0.8} 40.2 & \cellcolor[gray]{0.8} 69.8 \\
			\bottomrule
		\end{tabular}}
		\caption{\it \small Quantitative results of CSRNet+ enriched with a segmentation branch, on five popular datasets, against several approaches in the literature. ``pre-trained'' refers to models pre-trained on the large-scale UCF-QNRF dataset.\vspace{-9mm}}
		\label{table:SOTAresults}
	\end{table}

\subsection{Comparison with the state-of-the-art} \label{sec:SOTA}
\vspace{-1mm}
In the previous sections we evaluated the effect of reducing background mistakes for crowd counting by enriching a model with a segmentation branch (sec.~\ref{sec:ablation}). For completeness, we now compare this architecture against other works in the literature. Results are presented in table~\ref{table:SOTAresults}. Despite its simplicity, our approach achieves remarkably competitive performance on all the five datasets, on par with the state-of-the-art. 
We find these results very encouraging, as they show that sometimes there is no need for complex architectures, but rather for simple solutions that tackle the right problem. Finally, we present some qualitative results of this approach in fig.~\ref{Fig:results_regression_count}.

\vspace{-3mm}
\section{Conclusions} \label{sec:concl}
\vspace{-3mm}
We presented an extensive analysis on a problem that has been overlooked by the literature, yet it plays a fundamental part in the overall crowd counting performance.  
Our results showed that the problem of counting on background regions is significant and in it is responsible for 18-49\% of the total count error. Furthermore, we showed that this problem can be substantially mitigated by teaching the counting network the concept of background. By simply enriching a crowd counting network with a background segmentation branch we were able to reduce these mitakes by up to 83\%, leading to better crowd counting performance (up to 20\%). Finally, such a simple architectural change led to results on par with the state-of-the-art, on all the evaluated datasets. We find these results remarkable and a clear indication that future research should start addressing this problem more directly.

\clearpage
{
\bibliographystyle{splncs04}
\bibliography{egbib}

\begin{thebibliography}{10}
\providecommand{\url}[1]{\texttt{#1}}
\providecommand{\urlprefix}{URL }
\providecommand{\doi}[1]{https://doi.org/#1}

\bibitem{arteta16eccv}
Arteta, C., Lempitsky, V., Zisserman, A.: Counting in the wild. In: ECCV (2016)

\bibitem{boominathan16acm}
Boominathan, L., Kruthiventi, S.S., Babu, R.V.: Crowdnet: A deep convolutional
  network for dense crowd counting. In: ACM (2016)

\bibitem{cao18eccv}
Cao, X., Wang, Z., Zhao, Y., Su, F.: Scale aggregation network for accurate and
  efficient crowd counting. In: ECCV (2018)

\bibitem{chan08cvpr}
Chan, A.B., Liang, Z.S.J., Vasconcelos, N.: Privacy preserving crowd
  monitoring: Counting people without people models or tracking. In: CVPR
  (2008)

\bibitem{chan09cvpr}
Chan, A.B., Vasconcelos, N.: Bayesian poisson regression for crowd counting.
  In: CVPR (2009)

\bibitem{cheng19iccv}
Cheng, Z.Q., Li, J.X., Dai, Q., Wu, X., Hauptmann, A.: Learning spatial
  awareness to improve crowd counting. In: ICCV (2019)

\bibitem{guerrero15icpria}
Guerrero-G{\'o}mez-Olmedo, R., Torre-Jim{\'e}nez, B., L{\'o}pez-Sastre, R.,
  Maldonado-Basc{\'o}n, S., Onoro-Rubio, D.: Extremely overlapping vehicle
  counting. In: ibPRIA (2015)

\bibitem{idrees18eccv}
Idrees, H., Tayyab, M., Athrey, K., Zhang, D., Al-Maadeed, S., Rajpoot, N.,
  Shah, M.: Composition loss for counting, density map estimation and
  localization in dense crowds. In: ECCV (2018)

\bibitem{kang18bmvc}
Kang, D., Chan, A.: Crowd counting by adaptively fusing predictions from an
  image pyramid. In: BMVC (2018)

\bibitem{lempitsky10nips}
Lempitsky, V., Zisserman, A.: Learning to count objects in images. In: NIPS
  (2010)

\bibitem{li18cvpr}
Li, Y., Zhang, X., Chen, D.: Csrnet: Dilated convolutional neural networks for
  understanding the highly congested scenes. In: CVPR (2018)

\bibitem{liu18cvpr}
Liu, J., Gao, C., Meng, D., Hauptmann, A.G.: Decidenet: Counting varying
  density crowds through attention guided detection and density estimation. In:
  CVPR (2018)

\bibitem{liu19iccv}
Liu, L., Qiu, Z., Li, G., Liu, S., Ouyang, W., Lin, L.: Crowd counting with
  deep structured scale integration network. In: ICCV (2019)

\bibitem{ma19iccv}
Ma, Z., Wei, X., Hong, X., Gong, Y.: Bayesian loss for crowd count estimation
  with point supervision. In: ICCV (2019)

\bibitem{onoro16eccv}
Onoro-Rubio, D., L{\'o}pez-Sastre, R.J.: Towards perspective-free object
  counting with deep learning. In: ECCV (2016)

\bibitem{rodriguez11iccv}
Rodriguez, M., Laptev, I., Sivic, J., Audibert, J.Y.: Density-aware person
  detection and tracking in crowds. In: ICCV (2011)

\bibitem{ryan09dicta}
Ryan, D., Denman, S., Fookes, C., Sridharan, S.: Crowd counting using multiple
  local features. In: DICTA (2009)

\bibitem{sam18cvpr}
Sam, D.B., Sajjan, N.N., Venkatesh~Babu, R., Srinivasan, M.: Divide and grow:
  Capturing huge diversity in crowd images with incrementally growing cnn. In:
  CVPR (2018)

\bibitem{sam17cvpr}
Sam, D.B., Surya, S., Babu, R.V.: Switching convolutional neural network for
  crowd counting. In: CVPR (2017)

\bibitem{shi19cvpr}
Shi, M., Yang, Z., Xu, C., Chen, Q.: Revisiting perspective information for
  efficient crowd counting. In: CVPR (2019)

\bibitem{shi16cvpr}
Shi, W., Caballero, J., Husz{\'a}r, F., Totz, J., Aitken, A.P., Bishop, R.,
  Rueckert, D., Wang, Z.: Real-time single image and video super-resolution
  using an efficient sub-pixel convolutional neural network. In: CVPR (2016)

\bibitem{shi19iccv}
Shi, Z., Pascal, M., Snoek, C.G.M.: Counting with focus for free. In: ICCV
  (2019)

\bibitem{sindagi17iccv}
Sindagi, V.A., Patel, V.M.: Generating high-quality crowd density maps using
  contextual pyramid cnns. In: ICCV (2017)

\bibitem{sindagi19iccv}
Sindagi, V.A., Patel, V.M.: Multi-level bottom-top and top-bottom feature
  fusion for crowd counting. In: ICCV (2019)

\bibitem{ramavarior19arxiv}
Varior, R.R., Shuai, B., Tighe, J., Modolo, D.: Scale-aware attention network
  for crowd counting. In: arXiv preprint arXiv:1901.06026 (2019)

\bibitem{wan19cvpr}
Wan, J., Luo, W., Wu, B., Chan, A.B., Liu, W.: Residual regression with
  semantic prior for crowd counting. In: CVPR (2019)

\bibitem{wang11cvpr}
Wang, M., Wang, X.: Automatic adaptation of a generic pedestrian detector to a
  specific traffic scene. In: CVPR (2011)

\bibitem{wang19cvpr}
Wang, Q., Gao, J., Lin, W., Yuan, Y.: Learning from synthetic data for crowd
  counting in the wild. In: CVPR (2019)

\bibitem{wu05iccv}
Wu, B., Nevatia, R.: Detection of multiple, partially occluded humans in a
  single image by bayesian combination of edgelet part detectors. In: ICCV
  (2005)

\bibitem{xu19iccv}
Xu, C., Qiu, K., Fu, J., Bai, S., Xu, Y., Bai, X.: Learn to scale: Generating
  multipolar normalized density maps for crowd counting. In: ICCV (2019)

\bibitem{yan19iccv}
Yan, Z., Yuan, Y., Zuo, W., Tan, X., Wang, Y., Wen, S., Ding, E.:
  Perspective-guided convolution networks for crowd counting. In: ICCV (2019)

\bibitem{zhang9iccv}
Zhang, A., Yue, L., Shen, J., Zhu, F., Zhen, X., Cao, X., Shao, L.: Attentional
  neural fields for crowd counting. In: ICCV (2019)

\bibitem{zhang15cvpr}
Zhang, C., Li, H., Wang, X., Yang, X.: Cross-scene crowd counting via deep
  convolutional neural networks. In: CVPR (2015)

\bibitem{zhang18wacv}
Zhang, L., Shi, M.: Crowd counting via scale-adaptive convolutional neural
  network. In: WACV (2018)

\bibitem{zhang16cvpr}
Zhang, Y., Zhou, D., Chen, S., Gao, S., Ma, Y.: Single-image crowd counting via
  multi-column convolutional neural network. In: CVPR (2016)

\bibitem{zhou17cvprd}
Zhou, B., Zhao, H., Puig, X., Fidler, S., Barriuso, A., Torralba, A.: Scene
  parsing through ade20k dataset. In: CVPR (2017)

\end{thebibliography}
}

\end{document}